\documentclass[a4paper,fleqn,10pt]{cas-dc}
\usepackage{hyperref}
\usepackage{float}
\usepackage{setspace}
\usepackage{lipsum}
\usepackage{graphicx}
\usepackage{makecell}
\usepackage{color}
\usepackage{url}
\usepackage{amsfonts}
\usepackage{amsmath}
\usepackage{latexsym}
\usepackage[numbers]{natbib}

\xspaceaddexceptions{]}
\def\tsc#1{\csdef{#1}{\textsc{\lowercase{#1}}\xspace}}
\tsc{WGM}
\tsc{QE}
\tsc{EP}
\tsc{PMS}
\tsc{BEC}
\tsc{DE}

\makeatletter
\renewcommand\section{\@startsection{section}{1}{\z@}%
    {15pt \@plus 3\p@ \@minus 3\p@}%
    {4\p@}%
    {%\let\@hangfrom\relax
     \sectionfont\raggedright\hst[13pt]}}
\renewcommand\subsection{\@startsection{subsection}{2}{\z@}%
    {10pt \@plus 3\p@ \@minus 2\p@}%
    {.1\p@}%
    {%\let\@hangfrom\relax
     \ssectionfont\raggedright }}
\renewcommand\subsubsection{\@startsection{subsubsection}{3}{\z@}%
    {10pt \@plus 1\p@ \@minus .3\p@}%
    {.1\p@}%
    {%\let\@hangfrom\relax
     \sssectionfont\raggedright}}
\makeatother
     
\begin{document}
\let\WriteBookmarks\relax
\def\floatpagepagefraction{1}
\def\textpagefraction{.001}
\shorttitle{Biologically plausible deep learning -- but how far can we go with shallow networks?}
\shortauthors{Illing et al.}
%\begin{frontmatter}

\title [mode = title]{Biologically plausible deep learning -- but how far can we go with shallow networks?}                      

\author[1]{Bernd Illing}[]
\cormark[1]
\ead{bernd.illing@epfl.ch}
%\ead[url]{www.cvr.cc, cvr@sayahna.org}

%\credit{Conceptualization of this study, Methodology, Software}

\address[1]{School of Computer and Communication Science \& School of Life Science, EPFL, 1015 Lausanne, Switzerland}

\author[1]{Wulfram Gerstner}[]
\author[1]{Johanni Brea}[]

\cortext[cor1]{Corresponding author}

\begin{abstract}[S U M M A R Y]
Training deep neural networks with the error backpropagation algorithm %, however,
 is considered implausible from a biological perspective. 
 Numerous recent publications suggest elaborate models for biologically plausible variants of deep learning, typically defining success as reaching around 98\% test accuracy on the MNIST data set.
Here, we investigate how far we can go on digit (MNIST) and object (CIFAR10) classification with biologically plausible, local learning rules in a network with one hidden layer and a single readout layer. 
The hidden layer weights are either fixed (random or random Gabor filters) or trained with unsupervised methods (Principal/Independent Component Analysis or Sparse Coding) that can be implemented by local learning rules. 
The readout layer is trained with a supervised, local learning rule.
We first implement these models with rate neurons. % and systematically compare the classification performance between them and to networks trained with backpropagation. 
This comparison reveals, first, that unsupervised learning does not lead to better performance than fixed random projections or Gabor filters for large hidden layers. 
Second, networks with localized receptive fields perform significantly better than networks with all-to-all connectivity and can reach backpropagation performance on MNIST.
We then implement two of the networks - fixed, localized, random \& random Gabor filters in the hidden layer - with spiking leaky integrate-and-fire neurons and spike timing dependent plasticity to train the readout layer.  %after a transient phase of the pattern presentation.
These spiking models achieve > 98.2\% test accuracy on MNIST,  % and \textcolor{red}{48\%} test accuracy on CIFAR10.
which is close to the performance of %non-convolutional 
rate networks with one hidden layer trained with backpropagation.
The performance of our shallow network models is comparable to most current biologically plausible models of deep learning. 
Furthermore, our results with a shallow spiking network provide an important reference and suggest the use of datasets other than MNIST for testing the performance of future models of biologically plausible deep learning.

\end{abstract}

\maketitle

\section{Introduction}
\label{sec:introduction}

While learning a new task, synapses deep in the brain undergo task-relevant changes \citep{Hayashi-Takagi2015}.
These synapses are often many neurons downstream of sensors and many neurons upstream of actuators.
Since the rules that govern such changes deep in the brain are poorly understood, it is appealing to draw inspiration from deep artificial neural networks (DNNs) \citep{LeCun2015}.
% 1. Deep learning is great and state-of-the-art.
%DNNs achieve human-level performance on recognition and classification tasks;
% The success story of deep learning raises the question whether the brain solves such tasks in a similar way \citep{Marblestone2016a}.
% 2. Deep learning is not biologically plausible 
%both 
DNNs and the cerebral cortex share that information is processed in multiple layers of many neurons \citep{Yamins2016,Kriegeskorte2015} and that learning depends on changes of synaptic strengths \citep{Hebb1949}. 
However, learning rules in the brain are most likely different from the backpropagation algorithm \citep{Crick1989,Marblestone2016a,Whittington2019}.
Furthermore, biological neurons communicate by sending discrete spikes as opposed to real-valued numbers used in DNNs.
Differences like these suggest that there exist other, possibly nearly equally powerful, algorithms that are capable to solve the same tasks by using different, more biologically plausible mechanisms. 
Thus, an important question in computational neuroscience is how to explain the fascinating learning capabilities of the brain with biologically plausible network architectures and learning rules.
Moreover from a pure machine learning perspective there is increasing interest in neuron-like architectures with local learning rules, mainly motivated by the current advances in neuromorphic hardware \citep{Nawrocki2016}.\\
% 4. Usually tested on MNIST
Image recognition is a popular task to test the performance of neural networks.
Because of its relative simplicity and popularity, the MNIST dataset (28$\times$28-pixel grey level images of handwritten digits, \citet{LeCun}) is often used for benchmarking.
Typical performances of existing models are around 97-99\% classification accuracy on the MNIST test set (see \autoref{sec:relatedwork} and \autoref{tab:MNISTbenchmarks}).
Since the performances of many classical DNNs trained with backpropagation (but without data augmentation or convolutional layers, see table in \citet{LeCun}) also fall in this region, accuracies around these values are assumed to be an empirical signature of backpropagation-like deep learning \citep{Lillicrap2016,Sacramento2017,Tavanaei2018a,Whittington2019}.
It is noteworthy, however, that several of the most promising approaches that perform well on MNIST have been found to fail on harder tasks \citep{Bartunov2018} or at least need major modifications to scale to deeper networks \citep{Moskovitz2018}.\\
There are two obvious alternatives to supervised training of all layers with backpropagation. 
The first one is to fix weights in the first layer(s) at random values , as proposed by general approximation theory \citep{Barron1993} and the extreme learning field \citep{Huang2006a}. The second alternative is unsupervised training in the first layer(s). 
In both cases, only the weights of a readout layer are learned with supervised training.
Unsupervised methods are appealing since they can be implemented with local learning rules, see e.g. ``Oja's rule'' \citep{Oja1982,Sanger1989} for principal component analysis, nonlinear extensions for independent component analysis \citep{Hyvarinen1998} or algorithms in \citet{Olshausen1997,Rozell2008,Liu2012,Brito2016} for sparse coding.
A single readout layer can be implemented with a local rule as well. A candidate is the delta-rule (also called ``perceptron rule''), which may be implemented by pyramidal spiking neurons with dendritic prediction of somatic spiking \citep{Urbanczik14}. 
Since straightforward stacking of multiple fully connected layers of unsupervised learning does not reveal more complex features \cite{Olshausen1997} we focus here on networks with a single hidden layer (see also \citet{Krotov2019}).\\
The main objective of this study is to see how far we can go with networks with a single hidden layer and biologically plausible, local learning rules, preferably using spiking neurons. 
To do so we first compare the classification performance of different rate networks: networks trained with backpropagation, networks with fixed random projections or random Gabor filters in the hidden layer and networks where the hidden layer is trained with unsupervised methods (\autoref{subsec:rates}).
Since sparse connectivity is sometimes superior to dense connectivity \citep{Litwin-Kumar2017,Bartunov2018} and successful convolutional networks leverage local receptive fields, we investigate sparse connectivity between input and hidden layer, where each hidden neuron receives input only from a few neighboring pixels of the input image (\autoref{subsec:locrecfields}).
Finally we implement the simplest, yet promising and biologically plausible models - localized random projections and random Gabor filters - with spiking leaky integrate-and-fire neurons and spike timing dependent plasticity (\autoref{subsec:spikes}). We discuss the performance and implications of this simplistic model with respect to current models of biologically plausible deep learning.

\section{Related work}
\label{sec:relatedwork}

\begin{figure*}%[!t]
\centering
%\vspace{-5pt}
\includegraphics[width=0.97\textwidth]{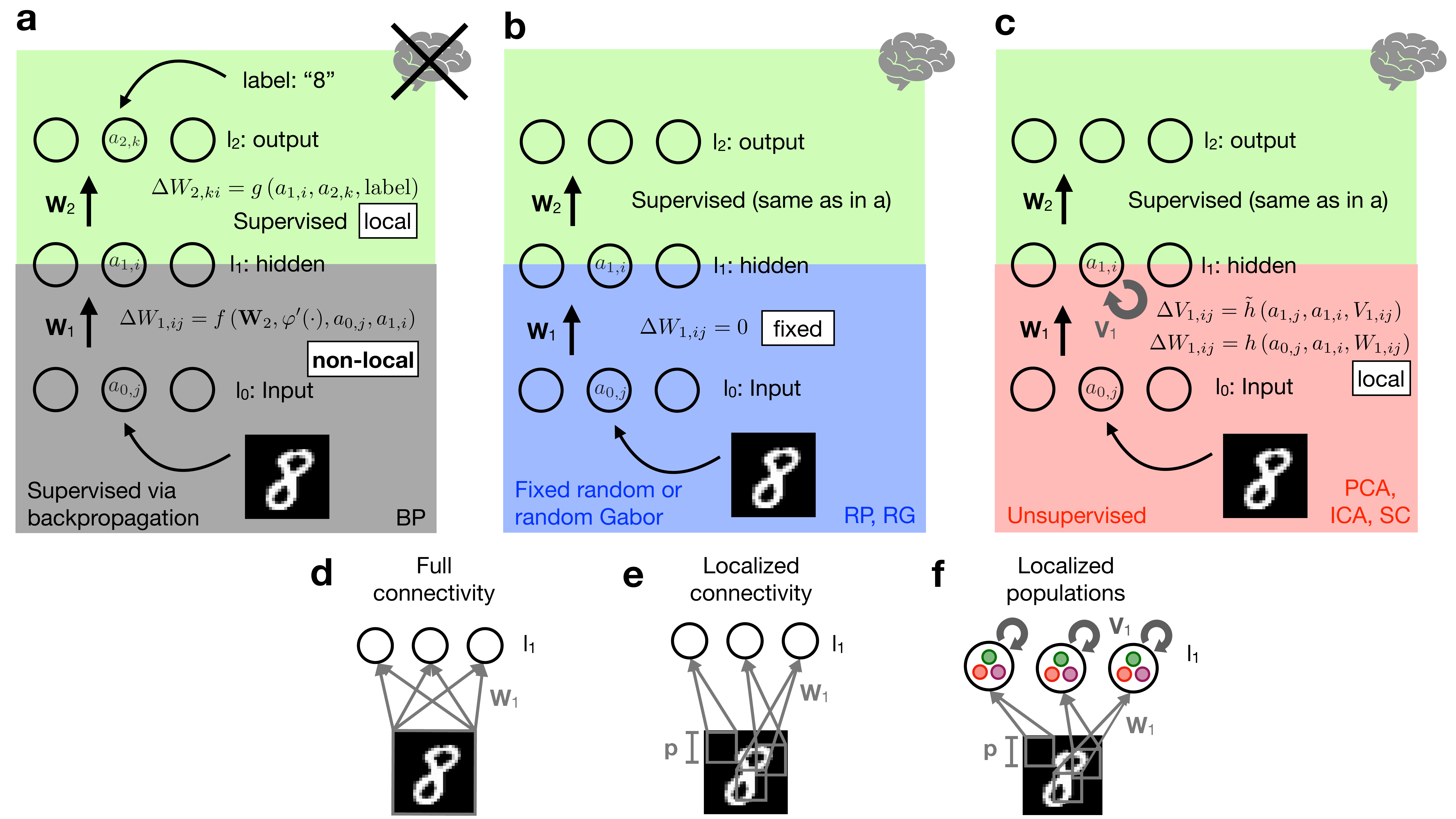}
%\vspace{-10pt}
\caption{The proposed network model has one hidden layer ($l_1$) and one readout layer ($l_2$) of nonlinear units (nonlinearity $\varphi(\cdot)$). Respective neural activations (e.g. $a_{0,j}$) and update rules (e.g. $\Delta W_{1,ij}$) are added. ($f(\cdot)$) $g(\cdot), h(\cdot)$ \& $\tilde{h}(\cdot)$ are (non-)local plasticity functions, i.e. using only variables (not) available at the synapse for the respective update. \textbf{a} Training with backpropagation (BP) through one hidden layer is biologically implausible since it is nonlocal (e.g. using $\textbf{W}_2$ \& $\varphi'(\cdot)$ from higher layers to update $\textbf{W}_1$, see \autoref{sec:BP}). \textbf{b} \& \textbf{c} Biologically plausible architecture with fixed Random Projections (RP) or fixed random Gabor filters (RG) (blue box in \textbf{b}) or unsupervised feature learning in the first layer (red box in \textbf{c}), and a supervised classifier in the readout layer $l_2$ (green boxes). All weight updates are local. $\textbf{W}$ stands for feed-forward, $\textbf{V}$ for recurrent, inhibitory weights. (Crossed out) brain icons in a,b \& c stand for (non-)bio-plausibility of the whole network. \textbf{d} \& \textbf{e} Illustration of fully connected and localized receptive fields of $\textbf{W}_1$. \textbf{f} For localized Principal/Independent Component Analysis ($l$-PCA/$l$-ICA) and Sparse Coding ($l$-SC) the hidden layer is composed of independent populations. Neurons within each population share the same localized receptive field and compete with each other while the populations are conditionally independent. For more model details, see \autoref{sec:methodsrate} - \autoref{sec:BP}.\label{fig:1}}
%\vspace{-10pt}
\end{figure*}

% 3. Numerous alternatives are currently explored
In recent years, many biologically plausible approaches to deep learning have been proposed, see e.g. \citet{Marblestone2016a,Whittington2019,Tavanaei2018a} for reviews.
Existing approaches usually use either involved architectures or elaborate mechanisms to approximate the backpropagation algorithm. Examples include the use of convolutional layers 
% mainly convolution (training with rates):
\citep{Tavanaei2016,Tavanaei2018a,Lee2018,Kheradpisheh2018} (and tables therein),
 dendritic computations
% mainly dendrites:
 \citep{Hussain2014,Guergiuev2016,Sacramento2017} 
 or backpropagation approximations such as feedback alignment
%  mainly backprop approx: 
\citep{Lillicrap2016,Baldi2016,Nokland2016,Samadi2017,Kohan2018,Bartunov2018} 
equilibrium propagation \citep{Scellier2017}, 
membrane potential based backpropagation \citep{Lee2016}, 
restricted Boltzmann machines and deep belief networks \citep{OConnor2013,Neftci2014}, 
(localized) difference target propagation \citep{Lee2015,Bartunov2018}, %7 layers! also 2% test error on MNIST!
 using reinforcement-signals \citep{Rombouts2015,Pozzi2018} or approaches using predictive coding \citep{Whittington2017}.
Many models implement spiking neurons to stress bio-plausibility 
%  spiking:  
 \citep{Liu2016,Neftci2017,Kulkarni2018,Wu2018,Liu2018,Tavanaei2018a} (and tables therein) or coding efficiency \citep{OConnor2017}. 
The conversion of DNNs to spiking neural networks (SNN) after training with backpropagation \citep{Diehl2015a} is a common technique to evade the difficulties of training with spikes. 
Furthermore, there are models including recurrent activity  
% recurrence: 
\citep{Spoerer2017, Bellec2018}, starting directly from realistic circuits 
% biologically plausible circuitry (moth circuit)
\citep{Delahunt2018}, or combining unsupervised and supervised training \citep{Krotov2019} as in this paper. 
We refer to \autoref{tab:MNISTbenchmarks} for an extensive list of current biologically plausible models tested on MNIST (see \autoref{tab:abbreviations} for abbreviations).

\begin{table*}[]
\centering

\caption{Alphabetical list of abbreviations in this paper.\label{tab:abbreviations}}
\resizebox{0.4\textwidth}{!}{
\begin{tabular}{|c|c|}
\hline
Abbreviation & Description\\\hline
AE & Autoencoder\\
ANN & Artificial Neural Network\\
BP & (Error-) Backpropagation\\
CNN / Conv. & Convolutional Neural Network\\
DBN & Deep Belief Network\\
DNN & Deep Neural Network\\
FA & Feedback Alignment\\
ICA & Independent Component Analysis\\
$l-$\dots & localized connectivity between input and hidden layer\\
LIF & Leaky Integrate-and-Fire\\
PCA & Principal Component Analysis\\
RBM & Restricted Boltzmann Machine\\
RG & Random Gabor filters\\
RL & Reinforcement Learning\\
RP & Random Projections\\
SC & Sparse Coding\\
SGD & Stochastic Gradient Descent\\
SNN & Spiking Neural Network\\
SP & Simple Perceptron\\ 
STDP & Spike Timing Dependent Plasticity\\ 
SVM & Support Vector Machine\\ \hline
\end{tabular}}

%\begin{scriptsize}
\caption{MNIST benchmarks for biologically plausible models of deep learning compared with models in this paper (\textbf{bold}). SNN: Spiking Neural Network, for other abbreviations see \autoref{sec:results}. Models are ranked by MNIST test accuracy (rightmost column). Parts of this table are taken from \citep{Tavanaei2018a,Kheradpisheh2018,Diehl2015}. Models using convolutional layers (CNN) are marked in \textcolor{orange}{orange}. See \autoref{tab:abbreviations} for abbreviations. 
%Note that the simple models in the $l$-RP class ($l$-RP, LIF rate \& spiking $l$-RP), marked in \textcolor{red}{red}, perform better than several more elaborate models. 
For conventional ANN/DNN/CNN MNIST benchmarks see \href{http://yann.lecun.com/exdb/mnist/}{table} in \citet{LeCun}.\label{tab:MNISTbenchmarks}}
\resizebox{0.98\textwidth}{!}{
\begin{tabular}{|c|c c c|c|}\hline
\textbf{Model} & \textbf{Neural coding} & \textbf{Learning type} & \textbf{Comments} & \textbf{Test accuracy (\%)}\\\hline
\textcolor{orange}{\makecell{Conv. SNN \citep{Wu2018}}} & Spikes & Supervised & \makecell{5 conv. layers, Spatio-Temporal BP} & \textcolor{orange}{99.3} \\ %\hline
\textcolor{orange}{\makecell{Conv. SNN \citep{Diehl2015a}}} & Rate & Supervised & \makecell{Conversion: rate $\to$ spike} & \textcolor{orange}{99.1} \\ %\hline
\textcolor{orange}{\makecell{Conv. Spiking AE\citep{Panda2016}}} & Spikes & Un/Supervised & \makecell{Stacked conv. AE with BP + sym. weights} & \textcolor{orange}{99.1} \\ %\hline
\textbf{$l$-RG} (this paper) & Rate & Un/Supervised & Only output layer learned &  \textbf{98.9} \\ %\hline
\textbf{$l$-BP} (this paper) & Rate & Supervised & \makecell{BP-benchmark of this paper} &  \textbf{98.8} \\ %\hline
\textbf{$l$-ICA} (this paper) & Rate & Un/Supervised & \makecell{ICs as features for SGD} & \textbf{98.8} \\ %\hline
\makecell{\textbf{$l$-FA} \citep{Bartunov2018} (\& this paper)} & Rate & Supervised  & \makecell{FA with localized rec. fields} & \textbf{98.7} \\ %\hline
\makecell{SNN \citep{Lee2016}} & Spikes & Supervised & \makecell{BP approx., weight symmetry} & 98.7 \\ %\hline
\makecell{\textbf{spiking LIF $l$-RG} (this paper)} & Spikes & Supervised & STDP (only output layer learned) & \textbf{98.6}\\ %\hline
\makecell{(Stoch.) Diff. Target Prop. \citep{Lee2015}} & Rate & Supervised & \makecell{Layer-wise AE, Target Prop.} & 98.5 \\ %\hline
%\textcolor{red}{\makecell{\textbf{LIF rate $l$-RP}\\(this paper)}} & Rate & Supervised & \makecell{Only output\\ layer learned} & \textbf{\textcolor{red}{98.5}}\\ %\hline
\makecell{Nonlin. Hebb + SGD \citep{Krotov2019}} & Rate & Un/Supervised  & \makecell{nonlin. Hebb + SGD (similar to this paper)} & 98.5\\ %\hline
\textbf{$l$-RP} (this paper) & Rate & Supervised & Only output layer learned &  \textbf{98.4} \\ %\hline
\textbf{$l$-SC} (this paper) & Rate & Un/Supervised & SC for 1. layer, SGD for 2. layer & \textbf{98.4} \\ %\hline
\textcolor{orange}{\makecell{Conv. SNN \citep{Kheradpisheh2018}}} & Spikes & Unsupervised & \makecell{3 Conv. layers, STDP, ext. SVM} & \textcolor{orange}{98.4}\\ %\hline 
\makecell{SNN \citep{OConnor2017}} & Pseudo-spike & Supervised & \makecell{Sparse, discrete activities, STDP} & 98.3 \\ %\hline
\makecell{Direct FA \citep{Nokland2016}} & Rate & Supervised & \makecell{Many hidden layers} & 98.3 \\ %\hline
\makecell{Spiking FA \citep{Lillicrap2016}} & Spikes & Supervised &  \makecell{3 hidden layers} & 98.2 \\ %\hline
\textbf{spiking LIF $l$-RP} (this paper) & Spikes & Supervised & STDP (only output layer learned) & \textbf{98.2}\\ %\hline
\textbf{$l$-PCA} (this paper) & Rate & Un/Supervised & \makecell{PCs as features for SGD} & \textbf{98.2} \\ %\hline
\makecell{Q-AGREL (RL-like) \citep{Pozzi2018}} & Rate & RL-like & \makecell{RL-like BP-approx.} & 98.2 \\ %\hline
\makecell{Forward propagation (FP) \citep{Kohan2018}} & Rate & Supervised &  \makecell{FP: BP approximation} & 98.1 \\ %\hline
\makecell{Spiking FA \citep{Neftci2017}} & Spikes & Supervised &\makecell{Direct FA} & 98 \\ %\hline
\makecell{Predictive coding \citep{Whittington2017}} & Rate & Supervised & \makecell{BP approx. by pred. coding} & 98 \\ %\hline
\textcolor{orange}{\makecell{Spiking CNN \citep{Tavanaei2016}}} & Rate/Spikes & Unsupervised & \makecell{Semi-online, STDP, ext. SVM} & \textcolor{orange}{98}\\ %\hline
%\textbf{FA-nd} (this paper) & Rate & Supervised & FA-nd & \makecell{FA without deriv.\\ of act. funct.} & \textbf{98.1} \\ %\hline
\makecell{Equilibrium Prop. \citep{Scellier2017}} & Rate & Supervised & \makecell{1 - 3 hidden layers} & 97 - 98\\ %\hline  
\makecell{Dendr. BP \citep{Sacramento2017}} & Spikes & Supervised & \makecell{Dendritic comp. for BP approx.} & 97.5 \\ %\hline
\makecell{Spiking FA \citep{Samadi2017}} & Spikes & Supervised &  \makecell{3 hidden layers} & 97 \\ %\hline
\makecell{Sparse/Skip FA \citep{Baldi2016}} & Rate & Supervised &  \makecell{Sparse- \& Skip-FA} & 96 - 97 \\ %\hline
\textcolor{orange}{\makecell{Spiking CNN \citep{Thiele2018}}} & Spikes & Unsupervised & \makecell{Recurrent Inhib., STDP} & \textcolor{orange}{96.6} \\ %\hline
\makecell{Spiking FA \citep{Guergiuev2016}} & Spikes & Supervised & \makecell{Dendritic comp. for BP approx.} & 96.3 \\ %\hline
\makecell{2 layer network \citep{Diehl2015}} & Spikes & Unsupervised & \makecell{Recurrent Inhib., purely unsuperv.} & 95\\ %\hline
\makecell{Spiking RBM/DBN \citep{OConnor2013}} & Rate & Supervised & \makecell{Conversion rate $\to$ spike} & 94.1\\ %\hline
\makecell{2 layer network \citep{Querlioz2013}} & Spikes & Unsupervised & \makecell{Memristive device} & 93.5 \\ %\hline
\textcolor{orange}{\makecell{Spiking HMAX/CNN \citep{Liu2018}}} & Spikes & Supervised & \makecell{STDP, HMAX preprocess.} & \textcolor{orange}{93} \\ %\hline
\makecell{Spiking RBM/DBN \citep{Neftci2014}} & Rate & Supervised & Neural sampling & 92.6 \\ %\hline
\makecell{Spiking RBM/DBN \citep{Neftci2014}} & Spikes & Supervised & Neural sampling & 91.9 \\ %\hline
\makecell{\textbf{SP} (this paper)} & Rate & Supervised & Direct classification on MNIST data & \textbf{91.9} \\ %\hline
\textcolor{orange}{\makecell{Spiking CNN \citep{Zhao2015}}} & Spike & Supervised & \makecell{Tempotron rule, sensor MNIST} & \textcolor{orange}{91.3}\\ %\hline
\makecell{Dendritic neurons \citep{Hussain2014}} & Rate & Supervised & \makecell{Nonlin. dendrites, neuromorphic appl.} & 90.3\\ \hline
\end{tabular}}

\end{table*}

\section{Results\label{sec:results}}

% In\autoref{fig:1} we illustrate the architecture of our network model. 
We study networks that consist of an input ($l_0$), one hidden ($l_1$) and an output-layer ($l_2$) of (nonlinear) units, connected by weight matrices $\textbf{W}_1$ and $\textbf{W}_2$ (\autoref{fig:1}).
%\autoref{fig:1}\textbf{A} shows supervised training with BP as the reference algorithm.
Training the hidden layer weights $\textbf{W}_1$ with standard supervised training involves (non-local) error backpropagation using summation over output units, the derivative of the units' nonlinearity ($\varphi'(\cdot)$) and the transposed weight matrix $\textbf{W}^T_2$ (\autoref{fig:1}a).
In the biologically plausible network considered in this paper (\autoref{fig:1}b \& c), the input-to-hidden weights $\textbf{W}_1$ are either fixed random, random Gabor filters or learned with an unsupervised method (Principal/ Independent Component Analysis or Sparse Coding). 
The unsupervised learning algorithms assume recurrent inhibitory weights $\textbf{V}_1$ between hidden units to implement competition, i.e. to make different hidden units learn different features. For more model details we refer to \autoref{sec:methodsrate} - \autoref{sec:BP}. \sloppy Code for all (rate \& spiking) models discussed below is publicly available at \href{https://github.com/EPFL-LCN/pub-illing2019-nnetworks}{https://github.com/EPFL-LCN/pub-illing2019-nnetworks}.
% Our approach is to first investigate rate models to figure out ways to train or initialize the first layer of our network (section \ref{subsec:rates}). The most promising method, fixed localized random projections ($l$-RP), is then implemented in a spiking LIF model (section \ref{subsec:spikes}). To efficiently find optimal parameters for the spiking model we use a LIF rate model, i.e. a rate model with similar dynamics and parameters as the spiking LIF model. Eventually the spiking LIF network is trained directly with STDP. For technical details and parameters we refer to the method section \ref{sec:methods}.

\subsection{Benchmarking biologically plausible rate models and backpropagation\label{subsec:rates}}

\begin{figure}%[!ht]
\centering
\vspace{-5pt}
%\hspace*{-1.5cm}
\includegraphics[width=0.45\textwidth]{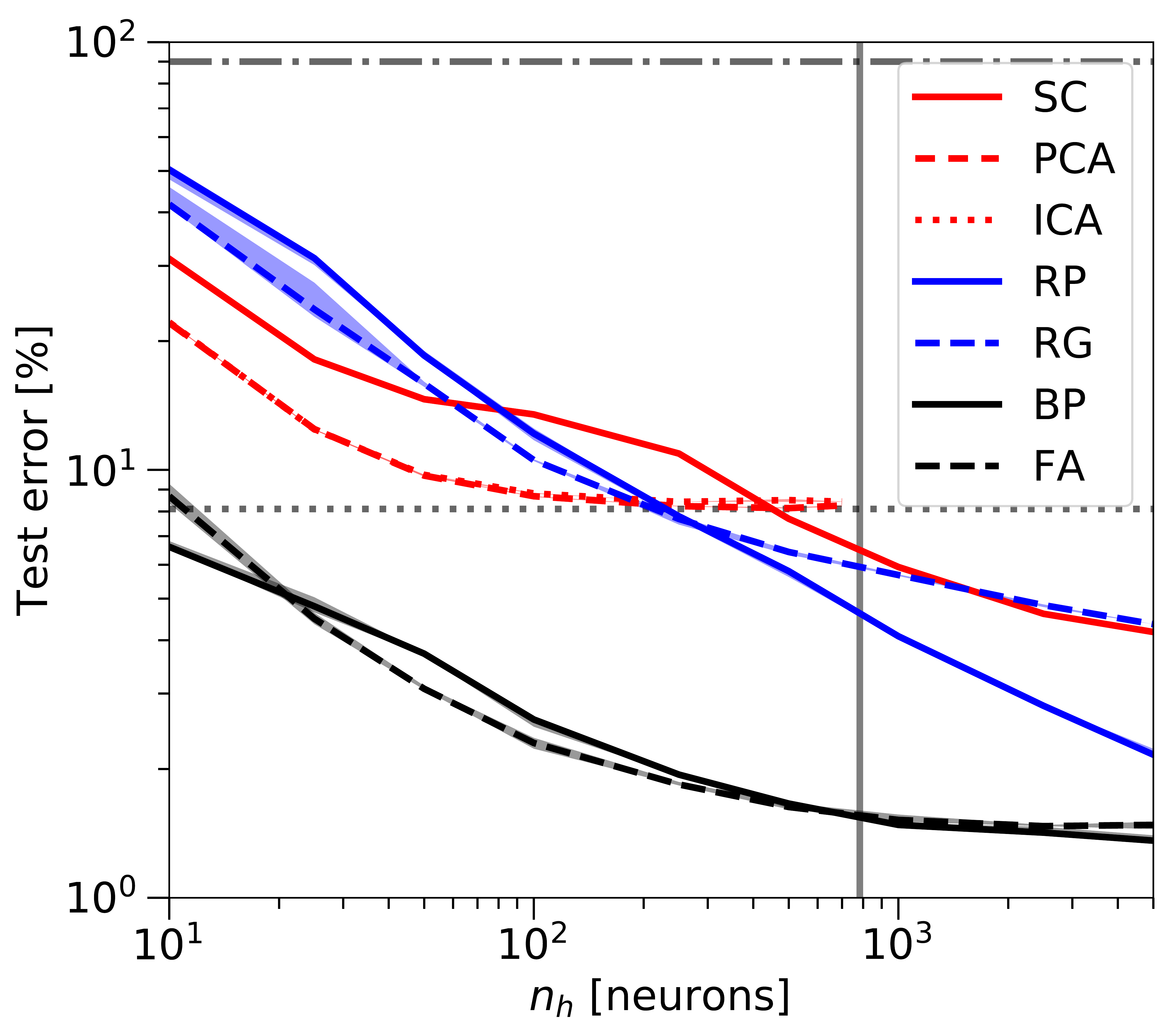}
%\vspace{-10pt}
\caption{MNIST classification with rate networks according to \autoref{fig:1}a-c with full connectivity (\autoref{fig:1}d). The test error decreases for increasing hidden layer size $n_h$ for all methods, i.e. Principal/Independent Component Analysis (PCA/ICA, curves are highly overlapping), Sparse Coding (SC), fixed Random Projections (RP) and fixed random Gabor filters (RG) as well as for the fully supervised reference algorithms Backpropagation (BP) and Feedback Alignment (FA). The dash-dotted line at 90 \% is chance level, the dotted line around 8 \% is the performance of a Simple Perceptron (SP) without hidden layer. The vertical line marks the input dimension $d = 784$, i.e. the transition from under- to overcomplete hidden representations. Note the log-log scale. \label{fig:2}}
\vspace{-10pt}
\end{figure}

To see how far we can go with a single hidden layer, we systematically investigate rate models using different methods to initialize or learn the hidden layer weights $\textbf{W}_1$ (see \autoref{fig:1} and methods \autoref{sec:methodsrate}-\autoref{sec:RPmethods} for details). 
We use two different ways to set the weights $\textbf{W}_1$ of the hidden layer: either using fixed Random Projections (\textbf{RP}) or Random Gabor filters (\textbf{RG}), see \autoref{fig:1}b \& blue curves in \autoref{fig:2}, or using one of the unsupervised methods Principal Component Analysis (\textbf{PCA}), Independent Component Analysis (\textbf{ICA}) or Sparse Coding (\textbf{SC}), see \autoref{fig:1}c \& red curves in \autoref{fig:2}.  
All these methods can be implemented with local, biologically plausible learning rules \citep{Oja1982,Hyvarinen1998,Olshausen1997}. We refer to the methods \autoref{sec:MethodsUnsupervised} for further details. 
As a reference, we train networks with the same architecture with standard backpropagation (\textbf{BP}, see \autoref{fig:1}a).
As a step from BP towards increased biologically plausibility, we include Feedback Alignment (\textbf{FA}, \citet{Lillicrap2016}) with fixed random feedback weights for error backpropagation (see methods \autoref{sec:BP} for further explanation). 
%Even though the random feedback weights in FA seem more biologically plausible than the symmetric feedback weights used in BP, FA still uses the derivative of the neural nonlinearity for error backpropagation. It appears biologically implausible that a real neuron can perform both computations, evaluating its nonlinearity in the forward- and the corresponding derivative in the backward path. To make FA even more realistic we add a version of it which omits the derivative of the activation function by setting it to 1 (\textbf{FA-nd}). 
A Simple Perceptron (\textbf{SP}) without a hidden layer serves as a further reference, %algorithm and lower performance bound
since it corresponds to direct classification of the input. 
We expect any biologically plausible learning algorithm to achieve results somewhere between SP (``lower'') and BP (``upper performance bound'')\\
%\textcolor{red}{TODO: Why did we exclude ICA? because of whitening issues. Hard to discuss. Suggest skipping it or only mention in text?!}
% For these reference algorithms, rectified linear units (ReLU) were used as nonlinearity.\\
The hidden-to-output weights $\textbf{W}_2$ are trained with standard stochastic gradient descent (SGD), using a one-hot representation of the class label as target.
Since no error backpropagation is needed for a single layer, the learning rule is local (``delta'' or ``perceptron''-rule). 
Therefore the two-layer network as a whole is biologically plausible in terms of online learning and synaptic updates using only local variables. 
For computational efficiency, we first train the hidden layer and then the output layer, however, both layers could be trained simultaneously.\\
% \subsubsection{Classification performance}
We compare the test errors on the MNIST digit recognition data set for varying numbers of hidden neurons $n_h$ (\autoref{fig:2}).
The PCA (red dashed) and ICA (red dotted) curves in \autoref{fig:2} end at the vertical line $n_h = d = 784$ because the number of principal/independent components (PCs/ICs), i.e. the number of hidden units $n_h$, is limited by the input dimension $d$.
Since the PCs span the subspace of highest variance, classification performance quickly improves when adding more PCs for small $n_h$ and then saturates for larger $n_h$. 
ICA does not seem to discover significantly more useful features than PCA, leading to similar classification performance.\\
SC (red solid line) extracts sparse representations that can be overcomplete ($n_h > d$), leading to a remarkable classification performance of around 96 \% test accuracy.
This suggests that the sparse representation and the features extracted by SC are indeed useful for classification, especially in the overcomplete case.\\
As expected, the performance of RP (blue solid) for small numbers of hidden units ($n_h < d$) is worse than for feature extractors like PCA, ICA or SC.
Also for large hidden layers, performance improves only slowly with $n_h$, which is in line with theory \citep{Barron1993} and findings in the extreme learning field \citep{Huang2006a}.
However, for large hidden layers sizes, RP outperforms SC.\\
As a reference, we also studied fixing the hidden layer weights to Gabor filters of random orientation, phase and size, located at the image center (RG, blue dashed, see \autoref{sec:RPmethods}). For hidden layers with more than 1000 neurons, SC is only marginally better than the network with fixed random Gabor filters.

%Initializing the hidden layer weights with random Gabor filters (RG) also allows for overcomplete representations. Filter parameters are chosen at random, but the functional form of the Gabor family puts a strong inductive bias onto the model. However, it turns out that Gabor filters spanning the whole input image lead to worse performance than SC and RP. This suggests the need for more local feature detectors that do not average out local information.\\
% This raises the question, if the SC features really help for classification or if the performance increase is just due to the high dimensionality of the hidden layer (as in RP).
% Part of the explanation might also be the saturation in accuracy for large $n_h$ since RP and SC seem to benefit much more from large hidden layer sizes than BP that saturates already around $n_h \approx 1000$.
For all tested methods and hidden layer sizes, performance is significantly worse than the one reached with BP (black solid in \autoref{fig:2}).
In line with \citep{Lillicrap2016}, we find that FA (black dashed) performs as well as BP on MNIST.
%However FA without the derivative of the activation function (FA-nd, purple curve) performs hardly better than RP. This clearly shows the limitations of this kind of top-down approximations of BP: if too many details are approximated or omitted, performance drops significantly.
Universal function approximation theory predicts lower bounds for the squared error that follow a power law with hidden layer size $n_h$  for both BP ($\mathcal{O}(1/n_h)$) and RP ($\mathcal{O}(1/n_h^{2/d})$, where $d$ is the input dimension \citep{Barron1994,Barron1993}). In the log-log-plot in \autoref{fig:2} this would correspond to a factor $d/2 = 784/2 = 392$ between the slopes of the curves of BP and RP, or at least a factor $d_{\mbox{eff}}/2 \approx 10$ using an effective dimensionality of MNIST (see methods \ref{sec:methodsrate}).
We find a much faster decay of classification error in RP and a smaller difference between RP and BP slopes than suggested by the theoretical lower bounds.\\
Taken together, these results show that the high dimensionality of the hidden layers is more important for reaching high performance than the global features extracted by PCA, ICA or SC. Tests on the object recognition task CIFAR10 lead to the same conclusion, indicating that this observation is not entirely task specific (see \autoref{subsec:locrecfields} for further analysis on CIFAR10).

\subsection{Localized receptive fields boost performance\label{subsec:locrecfields}}

\begin{figure*}%[!ht]
\centering
%\vspace{-40pt}
\includegraphics[width=0.85\textwidth]{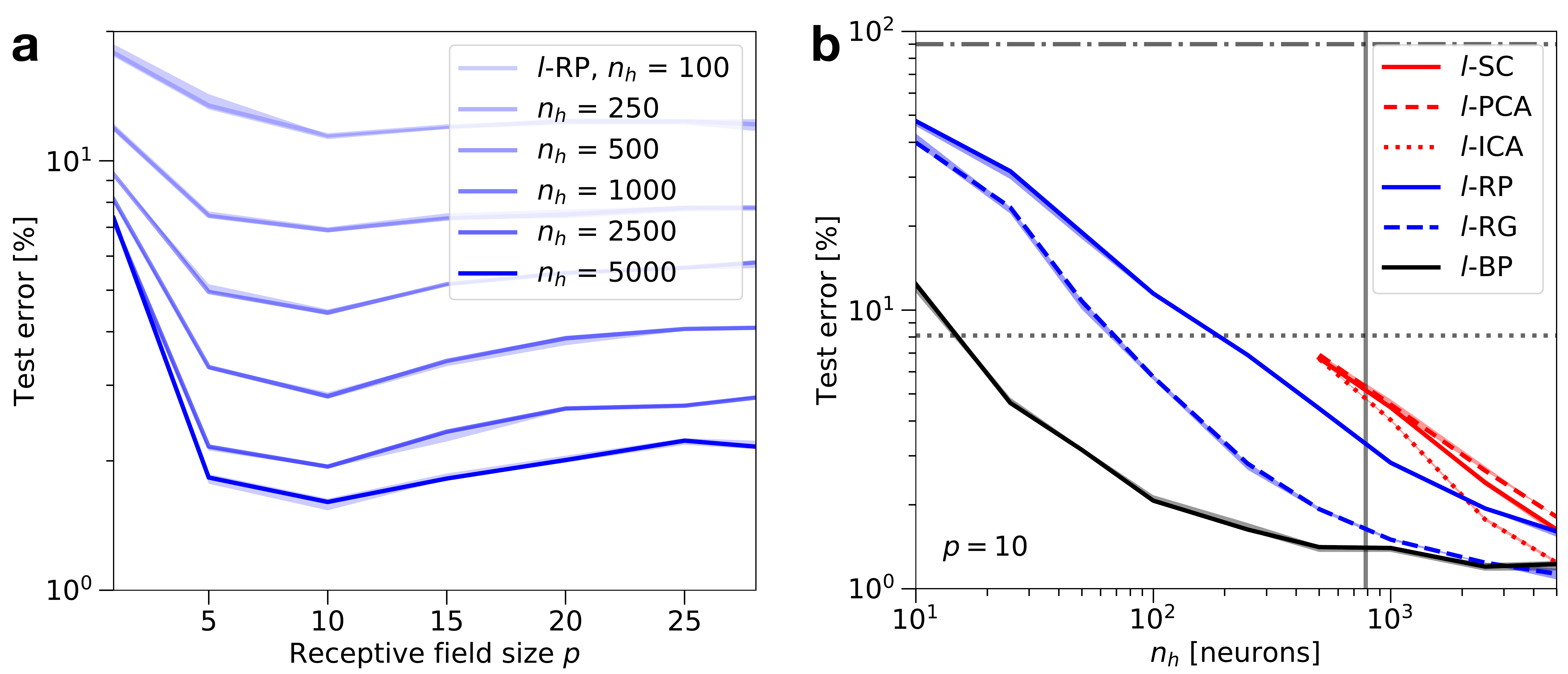}
%\vspace{-60pt}
\caption{Effect of localized connectivity on MNIST. \textbf{a} Test error for localized Random Projections ($l$-RP), dependent on receptive field size $p$ for different hidden layer sizes $n_h$. The optimum for receptive field size $p$ = 10 is more pronounced for large hidden layer sizes. Full connectivity is equivalent to $p$ = 28. Note the log-lin scale. \textbf{b} Localized receptive fields decrease test errors for all tested networks (compare \autoref{fig:2}): Principal/Independent Component Analysis ($l$-PCA/$l$-ICA), Sparse Coding ($l$-SC), Random Projections ($l$-RP), Random Gabor filters ($l$-RG) and Backpropagation ($l$-BP). The effect is most significant for $l$-ICA and $l$-RG, which approach $l$-BP performance for large $n_h$ and $p=10$, while all other methods reach test errors between $1 - 2\%$. All other reference lines as in \autoref{fig:2}. $l$-PCA/$l$-ICA \& $l$-SC use 500 independent populations in the hidden layer (see \autoref{fig:1}f) which constrains the hidden layer size to $n_h \geq 500$. Note the log-log scale.\label{fig:3}}
\vspace{-10pt}
\end{figure*}

There are good reasons to reduce the connectivity from all-to-all to localized receptive fields (\autoref{fig:1}e \& f): local connectivity patterns are observed in real neural circuits \citep{Hubel1962}, useful theoretically \citep{Litwin-Kumar2017} and empirically \citep{Bartunov2018}, and successfully used in convolutional neural networks (CNNs). Even though this modification seems well justified from both biological and algorithmic sides, it reduces the generality of the algorithm to input data such as images where neighborhood relations between pixels (i.e. input dimensions) are important.\\
To obtain localized receptive fields (called ``\textbf{$l$-}'' methods in the following) patches spanning $p\times p$ pixels in the input space are assigned to the hidden neurons. The centers of the patches are chosen at random positions in the input space, see \autoref{fig:1}e \& f. For localized Random Projections ($l$-RP) and localized random Gabor filters ($l$-RG) the weights within the patches are randomly drawn from the respective distribution and then fixed. 
For the localized unsupervised learning methods ($l$-PCA, $l$-ICA \& $l$-SC) the hidden layer is split into 500 independent populations. Neurons within each population compete with each other while different populations are independent, see \autoref{fig:1}f. This split implies a minimum number of $n_h = 500$ hidden neurons for these methods. For $l$-PCA and $l$-ICA a thresholding nonlinearity was added to the hidden layer to leverage the local structure (otherwise PCA/ICA act globally due to their linear nature, see methods \autoref{sec:MethodsUnsupervised}).\\
We test $l$-RP for different patch sizes $p$ and find an optimum around $p\approx 10$ (see \autoref{fig:3}a).
Note that $p = 1$ corresponds to resampling the data with random weights, and $p = 28$ recovers fully connected RP performance.
The other methods show similar optimal values around $p = 10$ (not shown).
The main finding here is the significant improvement in performance using localized receptive fields. 
All tested methods improve by a large margin when switching from full image to localized patches and some methods ($l$-RG and $l$-ICA) even reach BP performance for $n_h = 5000$ hidden neurons (see \autoref{fig:3}b). 
To achieve a fair comparison BP is also implemented with localized receptive fields ($l$-BP) which leads to a minor improvement compared to global BP. 
This makes local random projections or local unsupervised learning strong competitors to BP as biologically plausible algorithms in the regime of large, overcomplete hidden layers $n_h > d$ - at least for MNIST classification.\\
%We find that for $n_h = 100000$ hidden neurons $l$-BP and $l$-RP performance is not significantly different any more (both at 1.2\% test error for $p$ = 10) since $l$-BP performance saturates for large $n_h$ (see \autoref{fig:3}b).
% CIFAR10 results
To test whether localized receptive fields only work for the relatively simple MNIST data set (centered digits, uninformative margin pixels, no clutter, uniform features and perspective etc.) or generalizes to more difficult tasks, we apply it to the CIFAR10 data set \citep{Krizhevsky2013}.
We first reproduce a typical benchmark performance of a fully connected network with one hidden layer trained with standard BP ($\approx$ 56\% test accuracy, $n_h$ = 5000, see also \citet{Lin2016}).
Again, classification performance increases for increasing hidden layer size $n_h$ and localized receptive fields perform better than full connectivity for all methods.
Furthermore, as on MNIST, we can see similar performances for local feature learning methods ($l$-PCA, $l$-ICA \& $l$-SC) and local random features ($l$-RP, $l$-RG) in the case of large, overcomplete hidden layers (see \autoref{tab:CIFAR10results}).
% In \autoref{tab:CIFAR10results} we compare the performance of ($l$-)BP and ($l$-)RP on MNIST (see\autoref{fig:3}D) with the one on CIFAR10.
Also on CIFAR10, localized random filters and local feature learning reach the performance of biologically plausible models of deep learning \citep{Bartunov2018,Krotov2019} and come close to the performance of the reference algorithm $l$-BP. However, the difference remains statistically significant here.
%(last column of \autoref{tab:CIFAR10results}).
Given that the state-of-the-art performance on CIFAR10 with deep convolutional neural networks is close to 98\% (e.g. \citet{Real2018}), the limitations of our shallow local network and the well-known differences in difficulty between MNIST and CIFAR10 become apparent.\\
In summary, the main message of this section is that unsupervised methods, as well as random features, perform significantly better when applied locally. Equipped with local receptive fields our shallow network can outperform many current models of biologically plausible deep learning (see \autoref{tab:MNISTbenchmarks}). On MNIST some models ($l$-RG \& $l$-ICA) even reach backpropagation performance, while on CIFAR10 large differences to state-of-the-art deep convolutional networks remain.

\begin{table*}%[!ht]
\centering
\caption{Test accuracies (\%) on MNIST and CIFAR10 for rate networks and spiking LIF models. The Simple Perceptron (SP) is equivalent to direct classification on the data without hidden layer. All other methods use $n_h$ = 5000 hidden neurons and receptive field size $p$ = 10. Note that CIFAR10 has $d$ = 32$\times$32$\times$3 = 3072 input channels (the third factor is due to the color channels), MNIST only $d$ = 28$\times$28 = 784. The rate (spiking) models are trained for 167 (117) epochs. Best performing in bold. \label{tab:CIFAR10results}}
\begin{tabular}{|c c|c c c c c c|c|}
\hline
 \multicolumn{2}{|c|}{}& SP & $l$-PCA & $l$-ICA & $l$-SC & $l$-RP & $l$-RG & $l$-BP \\\hline %Factor $l$-RP/$l$-BP \\\hline\hline
\multirow{2}{*}{Rate} & CIFAR10 & 41.1 $\pm$ 0.1 & 50.8 $\pm$ 0.3 & 53.9 $\pm$ 0.3 & 50.2 $\pm$ 0.2 & 52.0 $\pm$ 0.4 & 55.6 $\pm$  0.2 & \textbf{58.3 $\pm$ 0.2} \\
 & MNIST & 91.9 $\pm$ 0.1 & 98.2 $\pm$ 0.02 & \textbf{98.8 $\pm$ 0.03} & 98.4 $\pm$ 0.07  & 98.4 $\pm$ 0.1 & \textbf{98.9 $\pm$ 0.05} & \textbf{98.8 $\pm$ 0.1} \\\hline
Spiking & MNIST & \multicolumn{4}{c}{-} & 98.2 $\pm$ 0.05 & 98.6 $\pm$ 0.1 & - \\\hline
\end{tabular}
\end{table*}

\subsection{Spiking localized random projections\label{subsec:spikes}}

Real neural circuits communicate with short electrical pulses, called spikes, instead of real numbers such as rates. We thus extend our shallow network model to networks of leaky integrate-and-fire (\textbf{LIF}) neurons.  
The network architecture is the same as in \autoref{fig:1}b. 
To keep it simple we implement the two models with fixed random weights with LIF neurons: fixed localized Random Projections (\textbf{$l$-RP}) and fixed localized random Gabor filters (\textbf{$l$-RG}) with patches of size $p\times p$ - as in \autoref{subsec:locrecfields}. 
The output layer weights $\textbf{W}_2$ are trained with a supervised spike timing dependent plasticity (\textbf{STDP}) rule.

\begin{figure*}%[!ht]
\centering
%\vspace{-5pt}
\includegraphics[width=\textwidth]{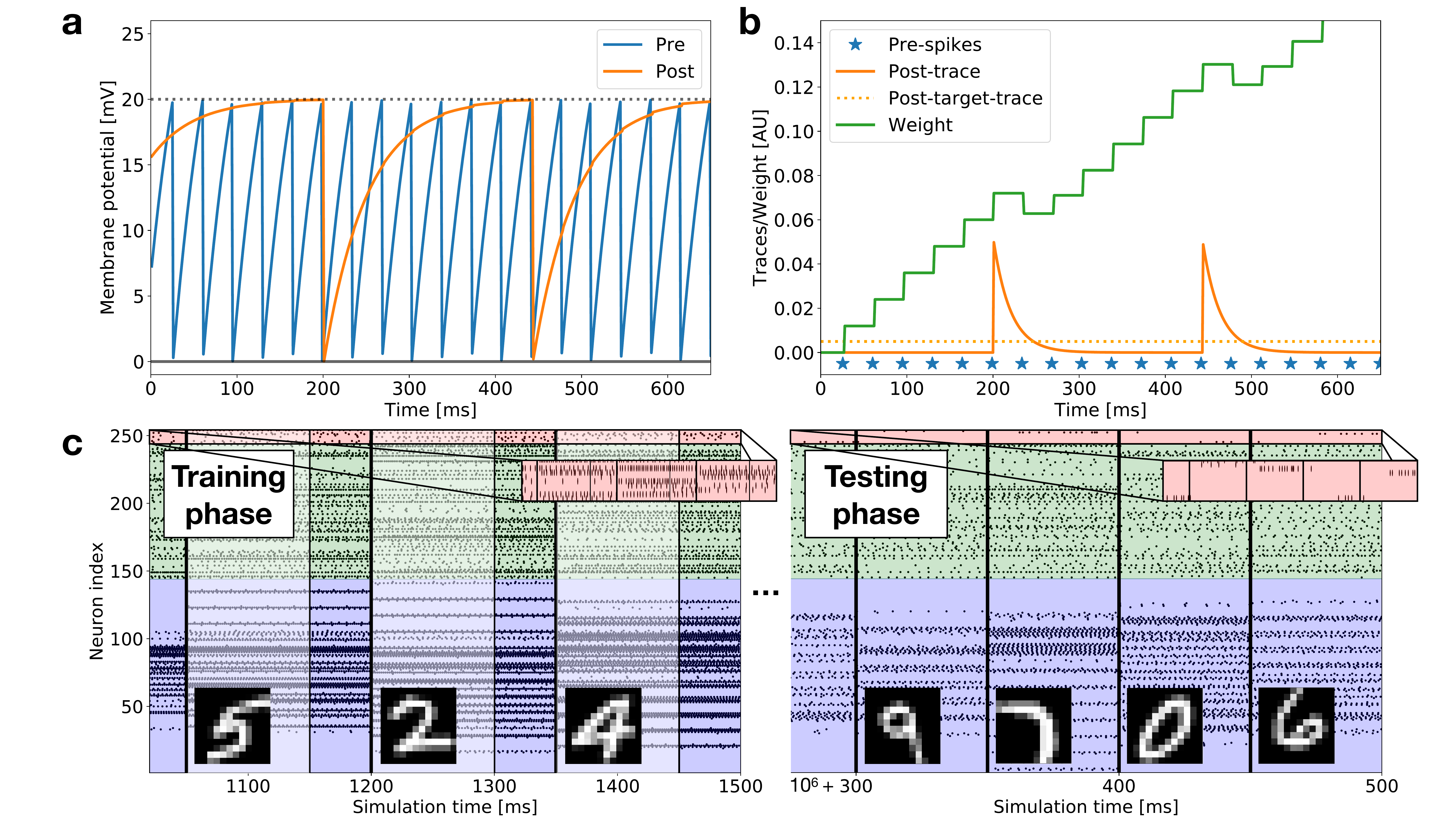}
\caption{Spiking LIF and STDP dynamics. \textbf{a} Dynamics of the pre- and postsynaptic membrane potentials, spike-traces and the weight value (\textbf{b}) of a toy example with two neurons and one interconnecting synapse. 
% Negative weight changes stem from the target-term (``delta-rule'') of the STDP rule (see methods \autoref{subsec:spiking}).
The weight decreases when the post-trace is above the post-target-trace (see \autoref{eq:STDPrule} and \autoref{subsec:spiking}).
Both neurons receive static supra-threshold external input: $I^{\mbox{ext}}_{\mbox{pre}} \gg I^{\mbox{ext}}_{\mbox{post}} \approx \vartheta$ (spiking threshold). Note that presynaptic spikes only slightly alter the postsynaptic potential since the weight is initially zero. \textbf{c} Rasterplot of a network trained on MNIST, where every spike is marked with a dot. The background color indicates the corresponding layers: input (blue, $n_0$ = 144 neurons), hidden (green, $n_1$ = $n_h$ = 100) and output (red, $n_2$ = 10). Bold vertical lines indicate pattern switches, thin lines indicate ends of transient phases (indicated by semi-transparency), during which learning is disabled. Left: Behaviour at the beginning of the training phase. Right: Testing period (learning off) after $6 \cdot 10^4$ presented patterns (1 epoch). As can be seen in the zoomed view of the 10 output layer neurons (red), the output layer has started to learn useful, 1-hot encoded class predictions. A downsampled ($12\times12$) version of MNIST is used for improved visibility. \label{fig:4}}
%\vspace{-10pt}
\end{figure*}

%\subsubsection{LIF and STDP dynamics} 
The spiking dynamics follow the usual LIF equations (see methods \autoref{subsec:spiking}) and the readout weights $\textbf{W}_2$ evolve according to a supervised delta rule via spike timing dependent plasticity (STDP) using post-synaptic spike-traces % $\mbox{tr}_j(t)$ \& 
$\mbox{tr}_i(t)$ and a post-synaptic target trace $\mbox{tgt}_i(t)$
\begin{eqnarray}\label{eq:STDPrule}
\tau_{\mbox{tr}} \frac{d \mbox{tr}_i(t)}{dt} &=& -\mbox{tr}_i(t) + \sum_f \delta\left(t - t_i^f\right)\\\nonumber
\Delta w_{2,ij} &=& \alpha \cdot \left(\mbox{tgt}^{\mbox{post}}_i(t) - \mbox{tr}^{\mbox{post}}_i(t)\right)\delta\left(t - t_j^f\right),
\end{eqnarray}
where $\alpha$ is the learning rate. 
Thus, for a specific readout weight $w_{2,ij}$, the post-synaptic trace is updated at every post-synaptic spike time $t_i^f$ and the weight is updated at every pre-synaptic spike time $t_j^f$.
The target trace is constant while a pattern is presented and uses a standard one-hot coding for the supervisor signal in the output layer ($l_2$).\\
To illustrate the LIF and STDP dynamics, a toy example consisting of one pre- connected to one post-synaptic neuron is integrated for 650 ms. The pre- and post-synaptic membrane potentials show periodic spiking (\autoref{fig:4}a) which induces post-synaptic spike traces and corresponding weight changes (\autoref{fig:4}b), according to \autoref{eq:STDPrule}. 
For the MNIST task, \autoref{fig:4}c shows a raster plot for an exemplary training and testing protocol. 
During activity transients after a switch from one pattern to the next, learning is disabled until regular spiking is recovered. 
We experienced that without disabling learning during these transient phases the networks never reached a low test error. 
This is not surprising, since in this phase the network activities carry information both about the previously presented pattern and the current one, but the learning rule is designed for network activities in response to a single input
pattern. 
It is also known that LIF neurons differ from biological neurons in response to step currents (see \citet{Naud2008} and references therein). 
During the testing period, learning is shut off permanently (see methods section \ref{subsec:spiking} for more details).
The LIF and STDP dynamics can be mapped to a rate model (see e.g. \cite{Diehl2015a} and \autoref{subsec:spiking} for details). However all following results are obtained with the fully spiking LIF/STDP model.

%\subsubsection{Classification results for spiking LIF $l$-RP}
When directly trained with the STDP rule of \autoref{eq:STDPrule}, the spiking LIF models closely approach the performance of their rate counterparts.
%The corresponding LIF rate model reaches 98.5\% test accuracy.
%Transferring weights learned with the LIF rate model into the spiking LIF model results in similar accuracies as achieved by the LIF rate model.
\autoref{tab:CIFAR10results} compares the performances of the rate and spiking LIF $l$-RP \& $l$-RG models with the reference algorithm $l$-BP (for same hidden layer size $n_h$ and patch size $p$, see \autoref{subsec:locrecfields}).
The remaining gap ($<$ 0.3\%) between rate model and spiking LIF model presumably stems from noise introduced by the spiking approximation of rates and the activity transients mentioned above. % and the shorter training time of the spiking model (only $6\cdot 10^6$ compared to $10^7$ iterations due to long simulation times). \\
Both, the rate and spiking LIF model of $l$-RP/$l$-RG achieve accuracies close to the backpropagation reference algorithm $l$-BP and fall in the range of performance of prominent, biologically plausible models, i.e. 98-99\% test accuracy (see \autoref{sec:relatedwork} and \autoref{tab:MNISTbenchmarks}).
Based on these numbers we conclude that the spiking LIF model of localized random projections using STDP is capable of learning the MNIST task to a level that is competitive with known benchmarks for spiking networks.

\section{Discussion\label{sec:discussion}}

%The rules that govern plasticity of synapses deep in the brain remain elusive.
In contrast to biologically plausible deep learning algorithms that are derived from approximations of the backpropagation algorithm \citep{Whittington2019,Lillicrap2016,Sacramento2017,Pozzi2018}, we focus here on shallow networks with only one hidden layer. 
The weights from the input to the hidden layer are either learned by unsupervised algorithms with local learning rules; or they are fixed. 
If fixed, they are drawn randomly or represent random Gabor filters.
The readout layer is trained with a supervised, local learning rule.\\
When applied globally, randomly initialized fixed weights/ Gabor filters (RP/RG) of large hidden layers lead to better classification performance than training them with unsupervised methods like Principal/Independent Component Analysis (PCA/ICA) or Sparse Coding (SC). Such observations also occur in different contexts, e.g. \citet{Dasgupta2018} showed that (sparse) random projections, combined with dimensionality expansion outperform known  algorithms for locality-sensitive hashing.
%This implies that the inductive bias of PCA and SC is not well suited for the task of digit classification and object recognition, however, better suited unsupervised methods might be found \citep{Krotov2019}. 
It may be interesting to search for alternative unsupervised, local learning rules with an inductive bias that is better adapted to image processing tasks than the one of SC.\\
Replacing all-to-all connectivity with localized input filters is such an inductive bias that already proved useful in supervised models \citep{Bartunov2018} but turns out to be particularly powerful in conjunction with unsupervised learning ($l$-PCA, $l$-ICA \& $l$-SC). Interestingly, non of the local unsupervised methods could significantly outperform localized random Gabor filters ($l$-RG). 
%Already for a hidden layer size of 5000 neurons the performance of $l$-RP almost reaches the performance of backpropagation on MNIST. 
Furthermore, we find that the performance scaling with the number of hidden units $n_h$ is orders of magnitudes better than the lower bound suggested by universal function approximation theory \citep{Barron1993}.\\
%, however, a different neuron nonlinearity is used there). 
To move closer to realistic neural circuits we implement our shallow, biologically plausible network with spiking neurons and spike timing dependent plasticity to train the readout layer. 
Spiking localized random projections ($l$-RP) and localized Gabor filters ($l$-RG) reach $>$98\% test accuracy on MNIST which lies within the range of current benchmarks for biologically plausible models for deep learning (see \autoref{sec:relatedwork} and \autoref{tab:MNISTbenchmarks}).
Our network model is particularly simple, i.e. it has only one trainable layer and does not depend on sophisticated architectural or algorithmic features typically necessary to approximate backpropagation \citep{Whittington2019}.
Instead it only relies on the properties of high-dimensional localized random projections.\\
Since we want to keep our models as simple as possible, we use online stochastic gradient descent (SGD, no mini-batches) with a constant learning rate. % in the non-spiking models. 
There are many known ways to further tweak the final performance, e.g. with adaptive learning rate schedules or data augmentation, but our goal here is to demonstrate that even a simple model with constant learning rate achieves results that are comparable with more elaborate approaches that use e.g. convolutional layers with weight sharing \citep{Panda2016}, backpropagation approximations \citep{Lee2016}, multiple hidden layers \citep{Lillicrap2016}, dendritic neurons \citep{Sacramento2017}, recurrence \citep{Diehl2015} or conversion from rate to spikes \citep{Diehl2015a}.\\
Above 98\% accuracy we also have to take into account a saturating effect of the network training: better models will only lead to subtle improvements in accuracy.
It is not obvious whether improvements are really a proof of having achieved deep learning or just the result of tweaking the models towards the peculiarities of the MNIST dataset.
Localized random filters or local unsupervised feature learning perform remarkably well compared to fully-connected backpropagation in shallow networks, even on more challenging data sets such as CIFAR10. This makes our model an important benchmark for future, biologically plausible models but also clearly highlights the limitations of our shallow two-layer model.  
A long time ago state-of-the-art deep learning has moved from MNIST to harder datasets, such as CIFAR10 or ImageNet \citep{Deng2009}.  
Yet MNIST seems to be the current reference task for most biologically plausible deep learning models (see \autoref{sec:relatedwork} and \autoref{tab:MNISTbenchmarks}).
We suggest that novel, progressive approaches to biologically plausible deep learning should significantly outperform the results presented here. 
Furthermore, they should be tested on tasks other than MNIST, where real deep learning capabilities become necessary.

\section{Acknowledgments}

This research was supported by the Swiss National Science Foundation (no. 200020\_165538 and 200020\_184615) and by the European Union Horizon 2020 Framework Program under grant agreement no. 785907 (HumanBrain Project, SGA2).

\appendix

\section{General rate model details}\label{sec:methodsrate}
%\subsection{Rate network model\label{subsec:methodsrate}}

We use a 3-layer (input $l_0$, hidden $l_1 = l_h$ and output $l_2$) feed-forward rate-based architecture with layer sizes ($n_0$ for input), $n_1$ (hidden) and $n_2$ (output, with $n_2$ = 10 = number of classes). The layers are connected via weight matrices $\textbf{W}_1 \in \mathbb{R}^{n_1 \times n_0}$ and $\textbf{W}_2 \in \mathbb{R}^{n_2 \times n_1}$ and each neuron receives bias from the bias vectors $\textbf{b}_1 \in \mathbb{R}^{n_1}$ and $\textbf{b}_2 \in \mathbb{R}^{n_2}$ respectively (see \autoref{fig:1}). The neurons themselves are nonlinear units with an element-wise, possibly layer-specific, nonlinearity $\textbf{a}_i = \varphi_l(\textbf{u}_i)$. The feed-forward pass of this model thus reads

\begin{eqnarray}\nonumber
\textbf{u}_{l+1} &=& \textbf{W}_{l+1} \textbf{a}_l + \textbf{b}_{l+1}\\
\textbf{a}_{l+1} &=& \varphi_{l+1} (\textbf{u}_{l+1}).
\end{eqnarray}

The Simple Perceptron (SP) only consists of one layer ($l_2$, $\textbf{W}_2 \in \mathbb{R}^{n_2 \times n_0}$, $\textbf{b}_2 \in \mathbb{R}^{n_2}$). The sparse coding (SC) model assumes recurrent inhibition within the hidden layer $l_1$. This inhibition is not modeled by an explicit inhibitory population, as required by Dale's principle \citep{Dale1935}, but direct, plastic, inhibitory synapses $\textbf{V}_1 \in \mathbb{R}^{n_1 \times n_1}$ are assumed between neurons in $l_1$. Classification error variances in \autoref{fig:2} \& \autoref{fig:3} are displayed as shaded, semi-transparent areas with the same colors as the corresponding curves. Their lower and upper bounds correspond to the 25\% and 75\% percentiles of at least 10 independent runs.\\
An effective dimensionality $d_{\mbox{eff}}$ of the MNIST data set can be obtained, e.g. via eigen-spectrum analysis, keeping 90\% of the variance. We obtain values around $d_{\mbox{eff}}\approx 20$. The measure proposed by \citet{Litwin-Kumar2017} gives the same value $d_{\mbox{eff}} \approx 20$. We checked that training a perceptron (1 hidden layer, $n_h$ = 1000, $10^7$ iterations, ReLU, standard BP) on the first 25 PCs of MNIST instead of the full data set leads to a comparable MNIST performance (1.7\% vs 1.5\% test error respectively).
Together, these findings suggest that the MNIST dataset lies mostly in a low-dimensional linear subspace with $d_{\mbox{eff}}\approx 25 \ll d$. 
The MNIST (\& CIFAR10) data was rescaled to values in [0,1] and mean centered, which means that the pixel-wise average over the data was subtracted from the pixel values of every image. Simulations were implemented and performed in the \href{https://julialang.org/}{\texttt{Julia}}-language. \sloppy The code for the implementation of our rate network models is publicly available at \href{https://github.com/EPFL-LCN/pub-illing2019-nnetworks}{https://github.com/EPFL-LCN/pub-illing2019-nnetworks}.

%\subsection{Unsupervised techniques}

%\subsubsection{Principal/Independent component analysis (PCA/ICA)\label{sec:PCAICAmethods}}
\section{Unsupervised methods (PCA, ICA \& SC)}\label{sec:MethodsUnsupervised}
In this paper we do not implement PCA/ICA learning explicitly as a neural learning algorithm but by a standard PCA/ICA algorithm (\href{https://github.com/JuliaStats/MultivariateStats.jl}{MultivariateStats.jl}) since biologically plausible online algorithms for both methods are well known \citep{Sanger1989,Hyvarinen1998}. For $d$-dimensional data such algorithms output the values of the $n \leq d$ first principal/ independent components as well as the corresponding subspace projection matrix $\textbf{P} \in \mathbb{R}^{n\times d}$. This matrix can directly be used as feedforward matrix $\textbf{W}_1$ in our network since the lines of $\textbf{P}$ correspond to the projections of the data onto the single/independent principal components. In other words each neuron in the hidden layer $l_1$ extracts another principal/independent component of the data. ICA was performed with the usual pre-whitening of the data.\\
Since PCA/ICA is a linear model, biases $\textbf{b}_1$ were set to \textbf{0} and $\varphi_1(\textbf{u}) = \textbf{u}$. With this, we can write the (trained) feed-forward pass of the first layer of our PCA/ICA model as follows:
\begin{eqnarray}
\textbf{a}_1 = \textbf{u}_1 = \textbf{W}_1\cdot \textbf{a}_0\mbox{~~~ with  }\textbf{W}_1 = \textbf{P}
\end{eqnarray}

Since the maximum number of principal/independent components that can be extracted is the dimensionality of the data, $n_{\mbox{max}} = d$, the number of neurons in the hidden layer $n_1$ is limited by $d$. This makes PCA/ICA unusable for overcomplete hidden representations as investigated for SC and RP. 
%Consistency between the used standard algorithm and neural implementations of PCA (``Sanger's'' rule \citet{Sanger1989}) was checked by comparing the extracted PCs and visualizing the learned projections (rows of $\textbf{P}$) for the case of 30 extracted PCs, i.e. $n = 30$.\\
In the localized version of PCA/ICA we assume the hidden layer to consist of independent populations, each extracting PCs/ICs of its respective localized receptive field (see \autoref{fig:1}). The hidden layer was divided into 500 of those populations, resulting in a minimum number of $n_h = 500$ hidden neurons (1 PC/IC per population) for these methods (and up to 10 PCs/ICs per population for $n_h = 5000$). The classifier was then trained on the combined activations of all populations of the hidden layer. Because PCA/ICA are linear methods the localized PCA/ICA version would not extract significantly different features unless we introduce a nonlinearity in the hidden units. This was done by simply thresholding the hidden activations (ReLU with threshold 0). No further optimization in terms of nonlinearity- and threshold-tuning was performed.\\

%\subsubsection{Sparse coding (SC)\label{sec:SCmethods}}
Sparse coding (SC) aims at finding a feature dictionary $\textbf{W}\in\mathbb{R}^{h \times d}$ (for $d$-dimensional data) that leads to an optimal representation $\textbf{a}_1 \in\mathbb{R}^h$ which is sparse, i.e. has as few non-zero elements as possible. The corresponding optimization problem reads:

\begin{eqnarray}\nonumber
\textbf{W}^{opt}, \textbf{a}_1^{opt} &=& \mbox{argmin }\mathcal{L}(\textbf{W},\textbf{a}_1)\\
\mathcal{L}(\textbf{W},\textbf{a}_1) &=& \frac{1}{2}\Vert \textbf{a}_0 - \textbf{W}^\top \textbf{a}_1 \Vert^2_2 + \lambda \Vert \textbf{a}_1\Vert_1.
\end{eqnarray}
Since this is a nonlinear optimization problem with latent variables (hidden layer) it cannot be solved directly. Usually an iterative two step procedure is applied (akin to the expectation-maximization algorithm) until convergence: First optimize with respect to the activities $\textbf{a}$ with fixed weights $\textbf{W}$. Second, assuming fixed activities, perform a gradient step w.r.t to weights.\\
We implement a biologically plausible SC model using a 2-layer network with recurrent inhibition and local plasticity rules similar to the one in \citet{Brito2016}. For a rigorous motivation (and derivation) that such a network architecture can indeed implement sparse coding we refer to \citet{Olshausen1997,Zylberberg2011a,Pehlevan2015,Brito2016}. We apply the above mentioned two step optimization procedure to solve the SC problem given our network model. The following two steps are repeated in alternation until convergence of the weights:

\begin{enumerate}

\item \textbf{Optimizing the hidden activations:}\\
We assume given and fixed weights $\textbf{W}_1$ and $\textbf{V}_1$ and ask for optimal hidden activations $\textbf{a}_1$. Because of the recurrent inhibition $\textbf{V}_1$ the resulting equation for the hidden activities $\textbf{a}_1$ is nonlinear and implicit. To solve this equation iteratively, we simulate the dynamics of a neural model with time-dependent internal and external variables $\textbf{u}_1(t)$ and $\textbf{a}_1(t)$ respectively. The dynamics of the system is then given by \citet{Zylberberg2011a,Brito2016}:
\begin{eqnarray}\label{eq:SCforward}\nonumber
\tau_u \frac{d\textbf{u}_1(t)}{dt} &=& - \textbf{u}_1(t) + \left(\textbf{W}_1\textbf{a}_0(t) - \textbf{V}_1 \textbf{a}_1(t)\right)\\
\textbf{a}_1(t) &=& \varphi(\textbf{u}_1(t))
\end{eqnarray}
In practice the dynamics is simulated for $N_{\mbox{iter}} = 50$ iterations, which leads to satisfying convergence (change in hidden activations $<$ 5\%).
%parameters 

\item \textbf{Optimizing the weights:}\\
Now the activities $\textbf{a}_1$ are kept fixed and we update the weights following the gradient of the loss function. The weight update rules are Hebbian-type local learning rules \citep{Brito2016}:
\begin{eqnarray}\label{eq:SCWABRITO}\nonumber
\Delta W_{1,ji} &=& \alpha_w \cdot  a_{0,i} \cdot a_{1,j}\\
\Delta V_{1,jk} &=& \alpha_v \cdot  a_{1,k} \cdot \left(a_{1,j} - \left\langle a_{1,j} \right\rangle\right)
\end{eqnarray}
$\left\langle \cdot \right\rangle$ is a moving average (low-pass filter) over several past hidden representations (after convergence of the recurrent dynamics) with some time constant $\tau_{\mbox{mav}}$, e.g. $\tau_{\mbox{mav}}$ = 100 patterns. 
At the beginning of the simulation (or after a new pattern presentation) $\tau_{\mbox{mav}}$ is increased starting from 0 to $\tau_{\mbox{mav}}$ during the first $\tau_{\mbox{mav}}$. 
The values of the rows of $\textbf{W}_1$ are normalized after each update, however this can also be achieved by adding a weight decay term. 
Additionally the values of $\textbf{V}_1$ are clamped to positive values after each update to ensure that the recurrent input is inhibitory. Also the diagonal of $\textbf{V}_1$ is kept at zero to avoid self-inhibition.

\end{enumerate}

During SC learning, at every iteration, the variables $\textbf{u}_1(t)$ and $\textbf{a}_1(t)$ are reset (to avoid transients) before an input is presented. 
Then for every of the $N$ iterations, \eqref{eq:SCforward} is iterated for $N_{\mbox{iter}}$ steps and the weights are updated according to \eqref{eq:SCWABRITO}.\\
% patchy implementation
Similar to localized PCA/ICA, the localized version of SC uses independent populations in the hidden layer (see \autoref{fig:1}). The SC algorithm above was applied to each population and its respective receptive field independently. The classifier was then trained on the combined activations of all populations of the hidden layer.
% For a detailed parameter list, see \autoref{tab:SC}.

\section{Fixed Random Filters (RP \& RG)\label{sec:RPmethods}}

For RP, the weight matrix $\textbf{W}_1$ between input and hidden layer is initialized randomly $\textbf{W}_1 \sim \mathcal{N}(0,\sigma^2)$ with variance-preserving scaling: $\sigma^2 \propto 1/n_0$. The biases $\textbf{b}_1$ are initialized by sampling from a uniform distribution $\mathcal{U}([0,0.1])$ between 0 and 0.1. In practice we used the specific initialization
\begin{eqnarray}\nonumber
\textbf{W}_1 &\sim & \mathcal{N}(0,\sigma^2) ~~\sigma^2 = \frac{1}{100~n_0} \\
\textbf{b}_1 &\sim & \mathcal{U}([0,0.1])
\end{eqnarray}
for RP (keeping weights fixed), SC, SP and also BP \& RF (both layers with $\textbf{W}_2, \textbf{b}_2$ and $n_1$ respectively).\\
For localized RP ($l$-RP), neurons in the hidden layer receive input only from a fraction of the input units called a receptive field. Receptive fields are chosen to form a compact patch over neighbouring pixels in the image space. For each hidden neuron a receptive field of size $p\times p$ ($p \in \mathbb{N}$) input neurons is created at a random position in the input space. The weight values for each receptive field (rf) and the biases are initialized as:
\begin{eqnarray}
\textbf{W}_{1,\mbox{rf}} &\sim & \mathcal{N}(0,\sigma_{\mbox{rf}}^2)~~~\sigma_{\mbox{rf}}^2 = \frac{c}{100~p}\\
\textbf{b}_1 &\sim & \mathcal{U}([0,0.1])
\end{eqnarray}
were the parameter $c = 3$ was found empirically through a grid-search optimization of classification performance. 
For exact parameter values, see \autoref{tab:RP}.\\
The (localized) random Gabor filters in RG have the same receptive field structure as in $l$-RP (see \autoref{sec:RPmethods}) but instead of choosing the weights within the receptive field as random values, they are choosen according to Gabor filters $\textbf{W}_1 \propto g(x,y)$. Here, $x$ and $y$ denote the pixel coordinates within the localized receptive field relative to the patch center. The Gabor filters have the following functional form:

\begin{eqnarray}
g\left(x,y; \lambda,\Theta,\psi,\sigma,\gamma\right) &=&\\\nonumber
\exp\left( -\frac{x'^2+\gamma^2 y'^2}{2\sigma^2}\right) &\cdot & \cos\left(2\pi\frac{x'}{\lambda}+\psi\right)\\\nonumber
\begin{pmatrix}
	x'\\y'
\end{pmatrix}
&=&
\begin{pmatrix}
    \cos\Theta & \sin\Theta  \\
    -\sin\Theta & \cos\Theta
\end{pmatrix}
\cdot
\begin{pmatrix}
	x\\ y
\end{pmatrix}
\end{eqnarray}

To obtain diverse, random receptive fields we draw the parameters $\lambda,\Theta,\psi,\sigma,\gamma $ of the Gabor functions from uniform distributions over some intervals. The bounds of the sampling interval are optimized using Bayesian optimization (\href{https://github.com/jbrea/BayesianOptimization.jl}{BayesianOptimization.jl}) with respect to classification accuracy on the training set.

\section{Classifier \& Supervised reference algorithms (BP, FA \& SP)}\label{sec:BP}
The connections $\textbf{W}_2$ from hidden to output layer are updated by a simple delta-rule which is equivalent to BP in a single-layer network and hence is biologically plausible. 
For having a reference for our biologically plausible models (\autoref{fig:1}b \& c), we compare it to networks with the same architecture (number of layers, neurons, connectivity) but trained in a fully supervised way with standard backpropagation (\autoref{fig:1}a). The forward pass of the model reads:
\begin{eqnarray}\nonumber
\textbf{u}_{l+1} &=& \textbf{W}_{l+1} \textbf{a}_l + \textbf{b}_{l+1}\\
\textbf{a}_{l+1} &=& \varphi_{l+1} (\textbf{u}_{l+1})
\end{eqnarray}

Given the one-hot encoded target activations $\textbf{tgt}$, the error $\tilde{\textbf{e}}_L$ is 
\begin{eqnarray}
\tilde{\textbf{e}}_L &=& \textbf{tgt} - \textbf{a}_L
\end{eqnarray} 
 when minimizing mean squared error (MSE)
\begin{eqnarray}
\mathcal{L}_{\mbox{MSE}} &=& \frac{1}{2}\Vert \textbf{tgt} - \textbf{a}_L \Vert_2^2
\end{eqnarray}
 or 
\begin{eqnarray}\nonumber
\textbf{p} &=& \mbox{softmax}\left(\textbf{a}_L\right)\\
\tilde{\textbf{e}}_L &=& \textbf{tgt} - \textbf{p}
\end{eqnarray} 
for the softmax/cross-entropy loss (CE),
\begin{eqnarray}\nonumber
\mathcal{L}_{\mbox{CE}} &=& - \sum_{i=1}^{n_L} \mbox{tgt}_i \cdot  \log\left(p_i\right).
\end{eqnarray}

Classification results (on the test set) for MSE- and CE-loss were found to be not significantly different. Rectified linear units (ReLU) were used as nonlinearity $\varphi (\textbf{u}_{l})$ for all layers (MSE-loss) or for the first layer only (CE-loss). \\
In BP the weight and bias update is obtained by stochastic gradient descent, i.e. $\Delta W_{l,ij} \propto \frac{\partial \mathcal{L}}{\partial W_{l,ij}}$. The full BP algorithm for deep networks reads \citep{Rumelhart1986}:
\begin{eqnarray}\label{eq:BP}\nonumber
\textbf{e}_L &=& \varphi '_L(\textbf{u}_{L}) \odot \tilde{\textbf{e}}_L\\\nonumber
\textbf{e}_{l-1} &=& \varphi '_{l-1}(\textbf{u}_{l}) \odot \textbf{W}_{l}^\top \textbf{e}_{l}\\\nonumber
\Delta\textbf{W}_l &=& \alpha\cdot\textbf{e}_l \otimes \textbf{a}_{l-1}  \\
\Delta\textbf{b}_l &=& \alpha\cdot\textbf{e}_l
\end{eqnarray}
% or equivalently,
% \begin{eqnarray}\label{eq:BP2}\nonumber
% \textbf{e}_L &=& \tilde{\textbf{e}}_L\\\nonumber
% \textbf{e}_{l-1} &=& \textbf{W}_{l}^\top \left(\varphi '_{l}(\textbf{u}_{l}) \odot \textbf{e}_{l}\right)\\\nonumber
% \Delta\textbf{W}_l &=& \alpha\cdot \left(\varphi '_{l}(\textbf{u}_{l}) \odot\textbf{e}_l\right) \otimes \textbf{a}_{l-1}  \\
% \Delta\textbf{b}_l &=& \alpha\cdot \varphi '_{l}(\textbf{u}_{l}) \odot \textbf{e}_l
% \end{eqnarray}
where $\odot$ stands for element-wise multiplication, $\otimes$ is the outer (dyadic) product, $\varphi '_l(\cdot )$ is the derivative of the nonlinearity and $\alpha$ is the learning rate. FA \citep{Lillicrap2016} uses a fixed random matrix $\textbf{R}_l$ instead of the transpose of the weight matrix $\textbf{W}_l^{\top}$ for the error backpropagation step in \eqref{eq:BP}. 
%An even more biologically plausible, but also more crude approximation of BP is FA-nd, which neglects the derivative of the activation function in FA, i.e. using a random feedback matrix  $\textbf{R}_l$ and setting $\varphi '_l(\textbf{u}_{l}) = \textbf{1}$ in \autoref{eq:BP}.\\

To allow for a fair comparison with $l$-RP, BP and FA were implemented with full connectivity and with localized receptive fields with the same initialization as in $l$-RP. During training with BP (or FA), the usual weight update \eqref{eq:BP} was applied to the weights within the receptive fields. The exact parameter values can be found in \autoref{tab:RP}.

%\subsection{Spiking implementation\label{subsec:spiking}}

%\subsubsection{LIF model}
\section{Spiking implementation of RP \& RG \label{subsec:spiking}}
The spiking simulations were performed with a custom-made event-based leaky integrate-and-fire (LIF) integrator written in the \href{https://julialang.org/}{\texttt{Julia}}-language. 
\sloppy Code is available at \href{https://github.com/EPFL-LCN/pub-illing2019-nnetworks}{https://github.com/EPFL-LCN/pub-illing2019-nnetworks}.
 For large network sizes, the exact, event-based integration can be inefficient due to a large frequency of events. 
We thus also added an Euler-forward integration mode to the framework. 
For sufficiently small time discretization (e.g. $\Delta t \leq 5\cdot 10^{-2}$ ms for the parameters given in \autoref{tab:LIF}) the error of Euler-forward integration does not have negative consequences on the learning outcome. 
The dynamics of the LIF network is given by:
\begin{eqnarray}\label{eq:LIF}\nonumber
\tau_m \frac{du_i(t)}{dt} &=& -u_i(t) + RI_i(t)\\\nonumber
\mbox{with~~} I_i(t) &=& I^{ff}_i(t) + I^{ext}_i(t) \\
&=& \sum_{j,f} w_{ij} \epsilon\left(t - t_j^f\right)+  I^{ext}_i(t)
\end{eqnarray}
and the spiking condition:
%\begin{eqnarray}
%\mbox{if 
$u_i(t) \geq \vartheta_i$: $u_i \to u_{\mbox{reset}}$
%}
%\end{eqnarray} 
, where $u_i(t)$ is the membrane potential, $\tau_m$ the membrane time-constant, $R$ the membrane resistance, $w_{ij}$ are the synaptic weights, $\epsilon(t) = \delta (t)/\tau_m$ is the post-synaptic potential evoked by a pre-synaptic spike arrival, $\vartheta_i$ is the spiking threshold and $u_{\mbox{reset}}$ the reset potential after a spike.\\
The input is split into a feed-forward ($I^{ff}(t)$) and an external ($ I^{ext}(t)$) contribution. Each neuron in the input layer $l_0$ ($n_0 = d$) receives only external input $I^{ext}$ proportional to one pixel value in the data. To avoid synchrony between the spikes of different neurons, the starting potentials and parameters (e.g. thresholds) for the different neurons are drawn from a (small) range around the respective mean values.\\
% learning rule
We implement STDP using %pre- and 
post-synaptic spike-traces % $\mbox{tr}_j(t)$ \& 
$\mbox{tr}_i(t)$ and a post-synaptic target-trace $\mbox{tgt}_i(t)$. 
\begin{eqnarray}\label{eq:spikelearning}
\tau_{\mbox{tr}} \frac{d \mbox{tr}_i(t)}{dt} &=& -\mbox{tr}_i(t) + \sum_f \delta\left(t - t_i^f\right)\\\nonumber%\textcolor{red}{CORRECT ?}\\
\Delta w_{ij} &=& g\left(\mbox{tr}^{\mbox{post}}_i(t),\mbox{tgt}^{\mbox{post}}_i(t)\right)\delta\left(t - t_j^f\right)
\end{eqnarray}
with the plasticity function 
\begin{equation}
g\left(\mbox{tr}^{\mbox{post}}_i(t),\mbox{tgt}_i(t)\right) = \alpha \cdot \left(\mbox{tgt}^{\mbox{post}}_i(t) - \mbox{tr}^{\mbox{post}}_i(t)\right).
\end{equation}
%The pre-synaptic trace does not appear explicitly since the update at every pre-synaptic spike implicitly includes the activity of the pre-synaptic neuron.\\
To train the network, we present patterns to the input layer and a target-trace to the output layer. The MNIST input is scaled by the input amplitude $\mbox{amp}_{\mbox{inp}}$, the targets $\textbf{tgt}(t)$ of the output layer are the one-hot-coded classes, scaled by the target amplitude $\mbox{amp}_{\mbox{tgt}}$. 
Additionally, every neuron receives a static bias input $I^{\mbox{ext}}_{\mbox{bias}} \approx \vartheta$ to avoid silent units in the hidden layer. 
Every pattern is presented as fixed input for a time $T_{\mbox{pat}}$ and the LIF dynamics as well as the learning evolves according to \eqref{eq:LIF} and \eqref{eq:spikelearning} respectively. Learning is disabled after pattern switches for a duration of $T_{\mbox{trans}} = 4\tau_m$ since the noise introduced by these transient phases was found to deteriorate learning progress.
With the parameters we used for the simulations (see \autoref{tab:LIF}), firing rates of single neurons in the whole network stayed below 1 kHz which was considered as a biologically plausible regime. 
For the toy example in \autoref{fig:4}a\& b we used static input and target with the parameters amp$_{\mbox{inp}}$ = 40, amp$_{\mbox{tgt}}$ = 5 (i.e. target trace = 0.005), $\vartheta_{\mbox{mean}}$ = 20,  $\sigma_{\vartheta}$= 0, $\tau_m$ = 50, $\alpha$ = $1.2\cdot 10^{-5}$. For the raster plot in \autoref{fig:4}c we used amp$_{\mbox{inp}}$ = 300, amp$_{\mbox{tgt}}$ = 300, $\vartheta_{\mbox{mean}}$ = 20,  $\sigma_{\vartheta}$= 0, $\tau_m$ = 50, $\alpha$ = $1.2\cdot 10^{-5}$, $T_{\mbox{pat}}$ = 50 ms, $T_{\mbox{trans}}$ = 100 ms. The LIF dynamics can be mapped to a rate model described by the following equations:
\begin{eqnarray}\nonumber
\textbf{u}_l &=& \textbf{W}_l \textbf{u}_{l-1} + R\textbf{I}^{ext}\\\nonumber
\textbf{a}_l &=& \varphi_{\mbox{LIF}}\left(\textbf{u}_l\right)\\
\Delta w_{ij} &=& \tilde{g}\left(a^{\mbox{pre}}_j,a^{\mbox{post}}_i,\mbox{tgt}^{\mbox{post}}_i\right) 
\end{eqnarray}
with the (element-wise) LIF-activation function $\varphi_{\mbox{LIF}}(\cdot)$ and the modified plasticity function $\tilde{g}(\cdot)$:
\begin{eqnarray}\nonumber
\varphi_{\mbox{LIF}}\left(u_k\right) &=& \left[\Delta_{\mbox{abs}} - \tau_m \ln\left(1 - \frac{\vartheta_k}{u_k}\right)\right]^{-1}\\\nonumber
\tilde{g}\left(a^{\mbox{pre}}_j,a^{\mbox{post}}_i,\mbox{tgt}^{\mbox{post}}_i\right) &=& \tilde{\alpha}\cdot a^{\mbox{pre}}_j \cdot \left(\mbox{tgt}^{\mbox{post}}_i - a^{\mbox{post}}_i\right)
\end{eqnarray}
The latter can be obtained by integrating the STDP rule of \autoref{eq:spikelearning} and taking the expectation over spike times. 
Most of the parameters of the spiking- and the LIF rate models can be mapped to each other directly (see \autoref{tab:LIF}). 
The learningrate $\alpha$ must be adapted since the LIF weight change depends on the presentation time of a pattern $T_{\mbox{pat}}$. In the limit of long pattern presentation times ($T_{\mbox{pat}} \gg \tau_m , \tau_{\mbox{tr}}$), the theoretical transition from the learning rate of the LIF rate model ($\tilde{\alpha}$) to the one of the spiking LIF model ($\alpha$) is 
%Attention: This is only true for the update TODO in the LIF-simulator!!!
%\begin{equation}
$\alpha = \frac{1000\mbox{ ms}}{T_{\mbox{pat}}\mbox{ [ms]}}\cdot 1000\cdot \tilde{\alpha}$,
%\label{eq:lr_conversion}
%\end{equation}

where the second factor comes from a unit change from Hz to kHz. It is also possible to train weight matrices computationally efficient in the LIF rate model and plug them into the spiking LIF model afterwards. 
The reasons for the remaining difference in performance presumably lie in transients and single-spike effects that cannot be captured by the rate model. 
Furthermore the new target was presented immediately after a pattern switch even though the activity obviously needs at least a couple time constants ($\tau_{\mbox{tr}}$ or $\tau_m$) to propagate through the network. 
Removing this asynchrony between input and target should further shrink the discrepancy between rate and spiking models.

\begin{table*}%[!hbt]

\section{Parameter tables\label{sec:parameters}}
For all simulations, we scaled the learning rate proportional to $1/n_h$ for $n_h > 5000$ to ensure convergence.

\centering
\caption{(Hyper-)Parameters for ($l$-) BP, FA, RP, RG (apart from weight initialization, see \autoref{sec:RPmethods}) \& SP as well as the supervised classifier on top of ($l$-) PCA, ICA and SC representations. Best performing parameters in bold. \label{tab:RP}}
\resizebox{0.8\textwidth}{!}{%
\begin{tabular}{|c|c|c|}
\hline
\textbf{Parameter} & \textbf{Description} & \textbf{Value}\\\hline
$n_h = n_1$ & Number of hidden units & [10,25,50,100,250,500,1000,2500,\textbf{5000}]\\%\hline
$p$ & Rec. field sizes (edge length) in units & [1,5,\textbf{10},15,20,25,28]\\%\hline
$\alpha_l$ & Learning rate & 1e-3\\%\hline
$N$ & Number of iterations & 1e7 ($\approx$ 167 epochs) \\%\hline
$\textbf{W}_{l}^{\text{init}}$ & Feed-forward weight initialization & $W_{l,ij}\sim\mathcal{N}(0,1)/(10\sqrt{n_{l-1}})$ \\%\hline
$\textbf{b}_{1}^{\text{init}}$ & Bias initialization & $b_{l,i}\sim\mathcal{U}\left([0,1]\right)/10$  \\
$\varphi_l (\cdot)$ & nonlinearity & ReLU \\
$N_{\text{pop}}$ & Number of populations in hidden layer ($l$-PCA, $l$-ICA \& $l$-SC) & [50,100,\textbf{500}] \\\hline
\end{tabular}}

\caption{(Hyper-)Parameters for SC. Best performing parameters in bold. \label{tab:SC}}
\resizebox{0.8\textwidth}{!}{
\begin{tabular}{|c|c|c|}
\hline
\textbf{Parameter} & \textbf{Description} & \textbf{Value}\\\hline
$n_h = n_1$ & Number of hidden units & [10,25,50,100,250,500,1000,2500,\textbf{5000}]\\%\hline
$p$ & Rec. field sizes (edge length) in units & [1,5,\textbf{10},15,20,25,28]\\%\hline
$\alpha_w$ & Learning rate for $\textbf{W}_1$ & 1e-3\\%\hline
$\alpha_v$ & Learning rate for $\textbf{V}_1$ & 1e-2\\%\hline
$\lambda$ & Sparsity parameter & [1e-4,1e-3,\textbf{1e-2},1e-1,1e-0]\\%\hline
$S$ & Resulting sparsity (fraction of 0-elements in $l_1$) & 90 - 99\% (dependent on $n_h$)\\%\hline
$\tau_{\text{mav}}$ & Time constant of the moving average & 1e-2 [1/patterns] \\%\hline
$\tau_u$ & Time constant of inner variable $\textbf{u}_1(t)$ & 1e-1 [1/iterations]\\%\hline
$N_{\text{iter}}$ & Number of iterations solving \autoref{eq:SCforward} & 50\\%\hline
$N$ & Number of iterations for SC & 1e5 \\%\hline
$\textbf{W}_{l}^{\text{init}}$ & Feed-forward weight initialization & $W_{l,ij}\sim\mathcal{N}(0,1)/(10\sqrt{n_{l-1}})$ \\%\hline
$\textbf{V}_{1}^{\text{init}}$ & Reccurent weight initialization & \textbf{0} \\%\hline
$\textbf{b}_{1}^{\text{init}}$ & Bias initialization & \textbf{0} (and kept fixed) \\%\hline
$\varphi_1 (\cdot)$ & nonlinearity of hidden SC units & ReLU $\text{max}(0,\cdot -\lambda)$  \\\hline
\end{tabular}}

\caption{(Hyper-)Parameters for the spiking LIF $l$-RP \& $l$-RG models (apart from weight initialization, \autoref{sec:RPmethods}). Input and target amplitudes are implausibly high due to the arbitrary convention $R$ = 1 $\Omega$. Best performing parameters in bold. \label{tab:LIF}}
\resizebox{0.8\textwidth}{!}{%
\begin{tabular}{|c|c|c|}
\hline
\textbf{Parameter} & \textbf{Description} & \textbf{Value}\\\hline
$n_h = n_1$ & Number of hidden units & [10,25,50,100,250,500,1000,2500,\textbf{5000}]\\%\hline
$p$ & Rec. field sizes (edge length) in units & [1,\textbf{10},28]\\%\hline
$\tau_m$ & Membrane time constant & 25 ms\\%\hline
$R$ & Membrane resistance & 1 $\Omega$\\%\hline
$\Delta_{\text{abs}}$ & Absolute refractory period & 0 ms\\%\hline
$\vartheta_i$ & Spiking thresholds & $\vartheta_i \sim\mathcal{N}\left(\vartheta_{\text{mean}},\sigma_{\vartheta}\right)$\\%\hline
$\vartheta_{\text{mean}}$ & Mean spiking threshold & 20 mV\\%\hline
$\sigma_{\vartheta}$ & Variance of spiking thresholds & 1 mV\\%\hline
$\text{amp}_{\text{inp}}$ & Input amplitude & 500 mA\\%\hline
$\text{amp}_{\text{tgt}}$ & Target amplitude & 500 mA\\%\hline 
$I^{\text{ext}}_{\text{bias}}$ & External bias input to all neurons & $\vartheta_{\text{mean}}$/R\\%\hline
$\tau_{\text{tr}}$ & Spike trace time constant & 20 ms\\%\hline
$u_{\text{reset}}$ & Reset potential & 0 mV \\%\hline
$\alpha$ & Learning rate & 2e-4 ($n_h$ = 5000, 5e-4 for Euler forward)\\%\hline
$\tilde{\alpha}$ & Learning rate for LIF rate model & 1e-8 (for $n_h$ = 5000)\\%\hline % 2.5e-10
$N$ & Number of iterations for spiking/rate model & 6e6/1e7 ($\approx$ 117/167 epochs)\\%\hline
$\textbf{W}_{l}^{\text{init}}$ & Feed-forward weight initialization & $W_{l,ij}\sim\mathcal{N}(0,1)\cdot 20/\sqrt{n_{l-1}}$ \\%\hline
$\tilde{\textbf{W}}_{l}^{\text{init}}$ & Feed-forward weight initialization (LIF rate) & $W_{l,ij}\sim\mathcal{N}(0,1)\cdot 20/\sqrt{n_{l-1}}$  \\%\hline
$T_{\text{pat}}$ & Duration of pattern presentation & 50 ms (train, 200 ms during testing) \\%\hline
$T_{\text{trans}}$ & Duration of the transient without learning & 100 ms\\%\hline
$\Delta t$ & Time step for Euler integrator & $\leq$ 5e-2 ms\\\hline
\end{tabular}}
\end{table*}

%% Take Care: Weight init and learning rate only true for CHANGES TO BE MADE in eventbaseintegrator framework!!!
%% Otherwise factor 1000/tau between spike and rate LIF models


%\clearpage

\bibliographystyle{unsrtnat}
\begin{footnotesize}
\bibliography{bib,lcnbibliography}  % name your BibTeX data base
\end{footnotesize}

%\newpage
%\clearpage
%\vfill\pagebreak

\makeatletter

\def\pct{\expandafter\@gobble\string\%}

\immediate\write\@auxout{\pct\space This is a test line.\pct }

\end{document}